\documentclass[11pt,a4paper]{article}

\pdfoutput=1 

\usepackage[hyperref]{acl2017}
\usepackage{times}
\usepackage{latexsym}
\usepackage{amsmath}
\usepackage{multirow}
\usepackage{enumitem}

\usepackage[utf8]{inputenc}
\usepackage{graphicx}
\usepackage{amsmath,amssymb,amsfonts}
\usepackage{booktabs}
\usepackage{csquotes}
\usepackage{color}
\usepackage{url}
\usepackage{tikz}
\usepackage{textcomp}

\usepackage{inconsolata}

\usepackage[belowskip=-2pt]{caption}
\usetikzlibrary{arrows}
\usetikzlibrary{positioning}
\usetikzlibrary{shadows}

\usepackage{environ}
\makeatletter
\newsavebox{\measure@tikzpicture}
\NewEnviron{scaletikzpicturetowidth}[1]{%
  \def\tikz@width{#1}%
  \begin{lrbox}{\measure@tikzpicture}%
  \BODY
  \end{lrbox}%
  \pgfmathparse{#1/\wd\measure@tikzpicture}%
  \BODY
}
\makeatother

\aclfinalcopy 

\newcommand{\ftextnumero}{{\fontfamily{txr}\selectfont \textnumero}}

\title{{EELECTION} at SemEval-2017 Task 10:\\ Ensemble of nEural Learners for kEyphrase ClassificaTION}

\author{Steffen Eger\textsuperscript{\textdagger\textdaggerdbl}, Erik-Lân Do Dinh\textsuperscript{\textdagger}, Ilia Kuznetsov\textsuperscript{\textdagger}, Masoud Kiaeeha\textsuperscript{\textdagger}, Iryna Gurevych\textsuperscript{\textdagger\textdaggerdbl}\\
\textsuperscript{\textdagger}Ubiquitous Knowledge Processing Lab (UKP-TUDA)\\
Department of Computer Science, Technische Universität Darmstadt\\
\textsuperscript{\textdaggerdbl}Ubiquitous Knowledge Processing Lab (UKP-DIPF)\\
German Institute for Educational Research and Educational Information\\
\url{http://www.ukp.tu-darmstadt.de}
}

\date{}

\begin{document}
\maketitle
\begin{abstract}
This paper describes our approach to the SemEval 2017 Task 10: \enquote{Extracting Keyphrases and Relations from Scientific Publications}, specifically to Subtask (B): \enquote{Classification of identified keyphrases}. 
We explored three 
different deep learning approaches: a character-level convolutional neural network (CNN),
a stacked learner with an MLP meta-classifier, and an attention based Bi-LSTM. 
From these 
approaches, we created an ensemble 
of differently hyper-parameterized systems, achieving a micro-$F_1$-score of 0.63 on the test data. Our approach ranks 2nd (score of 1st placed system: 0.64)
out of four according to this official score.
However, 
we 
erroneously
trained 2 out of 3 neural nets (the stacker and the CNN) on only roughly 15\% of the full data, namely, the original development set. When trained on the full data (training$+$development), our ensemble has a micro-$F_{1}$-score of 0.69. 
Our code is available from \url{https://github.com/UKPLab/semeval2017-scienceie}.
\end{abstract}

\section{Introduction}

Although scientific experiments are often accompanied by vast amounts of structured data, full-text scientific publications still remain one of the main means for communicating academic knowledge. Given the dynamic nature of modern research 
and its ever-accelerating pace, 
it is crucial to automatically analyze new works 
in order to have a complete 
picture of advances in a given field.

Recently, some progress has been made in this direction for the fixed-domain use case\footnote{ E.g. BioNLP: \url{http://2016.bionlp-st.org/}}. However, creating a universal open-domain system still remains a challenge due to significant domain differences between articles originating from different fields of research. The \mbox{SemEval~2017~Task~10}: ScienceIE \cite{augenstein2017scienceie} promotes the multi-domain use case, providing source articles from three domains: Computer Science, Material Sciences and Physics. The task 
consists of three subtasks, namely 
(A) identification of keyphrases, (B) classifying them into broad domain-independent classes and (C) inferring relations between the identified keyphrases. 

For example, for the input sentence `The thermodynamics of copper-zinc alloys (brass) was subject of numerous investigations' the following output would be expected:






\begin{enumerate}[label=(\Alph*),itemsep=-0.3em,topsep=0.7em]
\item 
  \begin{enumerate}[label=\arabic*.,noitemsep,nolistsep]
    \addtolength{\itemindent}{-0.9em}
    \item The thermodynamics of copper-zinc alloys
    \item copper-zinc alloys
    \item brass
  \end{enumerate}
\item 
  \begin{enumerate}[label=\arabic*.,noitemsep,nolistsep]
    \addtolength{\itemindent}{-0.9em}
    \item \texttt{TASK}
    \item \texttt{MATERIAL}
    \item \texttt{MATERIAL}
  \end{enumerate}
\item 
  \begin{enumerate}[label={},noitemsep,nolistsep]
    \addtolength{\itemindent}{-0.9em}
    \item \texttt{synonym(2,3)}
  \end{enumerate}
\end{enumerate}

Our submission focuses on (B) keyphrase classification given 
item boundaries. We 
avoid task-specific feature engineering, which would potentially render the system domain-dependent. Instead, 
we build
an ensemble of several deep learning classifiers
detailed in \S\ref{sec:approaches}, whose inputs are word embeddings learned from general domains.

\label{sec:intro}
\section{Task and Data}
In the annotation scheme proposed by the task organizers, keyphrases denoting a scientific model, algorithm or process should be classified as \textit{\textbf{P}ROCESS} (P), which also comprises methods (e.g.\ `backpropagation'), physical equipment (e.g.\ `plasmatic nanosensors', `electron microscope') and tools (e.g.\ `MATLAB').
\textit{\textbf{T}ASK} (T) contains concrete research tasks (e.g.\ `powder processing', `dependency parsing') and research areas (e.g.\ `machine learning')
, while \textit{\textbf{M}ATERIAL} (M)
includes
physical materials (e.g.\ `iron', `nanotube'), 
and
corpora or datasets (e.g.\ `the CoNLL-2003 NER corpus').

The corpus for the shared task consisted of 500 
journal articles retrieved from \mbox{ScienceDirect}\footnote{ \url{http://www.sciencedirect.com/}}, evenly distributed among Computer Science, Material Sciences and Physics domains.
It was split into three segments of 350 (training), 50 (development), and 100 (test) documents.
The corpus used in subtask (B) contains paragraphs of those articles, annotated with spans of keyphrases. Table~\ref{table:stats} shows the  distribution of the classes M, T, and P in the data. 
We note that class T is underrepresented and makes up less than 16\% of all instances. 

\begin{table}[!htb]
  \centering
  \begin{tabular}{lrrr} \hline\toprule
    & Material & Process & Task \\ \midrule
    Train+Dev & 40\% & 44\% & 16\% \\
    Test & 44\% & 47\% & 9\%\\
    \bottomrule
  \end{tabular}
  \caption{Class distribution in the datasets.}
  \label{table:stats}
\end{table}

Inter-annotator agreement for the dataset was 
published to be between 0.45 and 0.85 (Cohen's $\kappa$) \cite{augenstein2017scienceie}. Reviewing similar annotation efforts \cite{Qasemizadeh2016} already shows that despite the seemingly simple annotation task,
usually annotators do not reach 
high
agreement neither on span of annotations nor the class assigned to each span\footnote{$F_{1}$-scores ranging from 0.528 to 0.755 for span boundaries and from 0.471 to 0.635 for semantic categories.}.
\label{sec:data}
\section{Implemented Approaches}

In this section, we describe the individual systems that form the basis of our experiments 
(see \S\ref{sec:submitted}).

Our basic setup for all of our systems was as follows. For each keyphrase we extracted its \textbf{left context}, \textbf{right context} and the {keyphrase} itself (\textbf{center}). We represent each of the three contexts as the \emph{concatenation} of their word tokens: to have fixed-size representations, we limit the left context to the $\ell$ previous tokens, the right context to the $r$ following tokens and the center to the $c$ initial tokens of the keyphrase. We consider $\ell,r$ and $c$ as hyper-parameters of our modeling. If necessary, we pad up each respective context with `empty' word tokens. We then map each token 
to a $d$-dimensional word embedding. The choices for word embeddings are described below. To summarize, we frame our classification problem as 
a mapping $f_\theta$ ($\theta$ represents model parameters) from concatenated word embeddings to one of the three classes \textit{MATERIAL}, \textit{PROCESS}, and \mbox{\textit{TASK}}:
\begin{align*}
  f_\theta:\mathbb{R}^{\ell\cdot d}\times \mathbb{R}^{c\cdot d}\times \mathbb{R}^{r\cdot d}\rightarrow \{\text{M},\text{P},\text{T}\}.
\end{align*}
Next, we describe the embeddings that we used and subsequently the machine learning models $f_\theta$. 

\subsection*{Word Embeddings}
We experimented with three 
kinds of 
word embeddings. 
We use the popular Glove embeddings \cite{Pennington:2014} (6B) of dimensions 50, 100, and 300, which largely capture semantic information. 
Further we employ the more syntactically oriented 300-dimensional embeddings of
\citet{Levy:2014}, 
as well as the 
300-dimensional embeddings of \citet{Komninos:2016}, which 
are trained to predict both dependency- 
and
standard window-based context. 

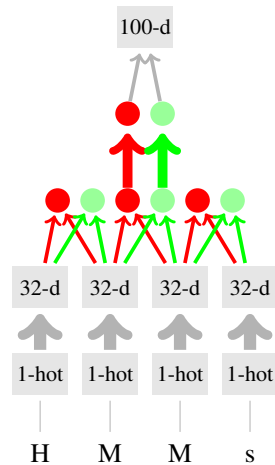
\begin{figure}[!b]
  \centering
  \resizebox{0.25\textwidth}{!}{%
    \def\layersep{1.25cm}


\begin{tikzpicture}[shorten >=1pt,->,draw=black!30, node distance=\layersep,transform shape,rotate=90]  
  \tikzstyle{every pin edge}=[-,shorten <=1pt]
  \tikzstyle{neuron}=[circle,fill=black!25,minimum size=10pt,inner sep=0pt]
  \tikzstyle{neuron2}=[circle,fill=green!40,minimum size=10pt,inner sep=0pt]
    \tikzstyle{neuronbox}=[rectangle,fill=black!10,minimum size=17pt,inner sep=0pt,line width=1mm];
    \tikzstyle{input neuron}=[neuron, fill=green!50];
    \tikzstyle{output neuron}=[neuron, fill=red!50];
    \tikzstyle{hidden neuron}=[neuron, fill=blue!50];
    \tikzstyle{annot} = [text width=4em, text centered]
    \tikzset{hoz/.style={rotate=-90}}   

    \node[neuronbox, pin=left:\rotatebox{-90}{\parbox[t][][r]{8mm}{\centering H}}] (H1) at (0,-5) {\rotatebox{-90}{\parbox[t][][r]{8mm}{\centering {\small 1-hot}}}};
    \node[neuronbox, pin=left:\rotatebox{-90}{\parbox[t][][r]{8mm}{\centering M}}] (M2) at (0,-6) {\rotatebox{-90}{\parbox[t][][r]{8mm}{\centering {\small 1-hot}}}};
    \node[neuronbox, pin=left:\rotatebox{-90}{\parbox[t][][r]{8mm}{\centering M}}] (M3) at (0,-7) {\rotatebox{-90}{\parbox[t][][r]{8mm}{\centering {\small 1-hot}}}};
    \node[neuronbox, pin=left:\rotatebox{-90}{\parbox[t][][r]{8mm}{\centering \textcolor{white}{S}s\textcolor{white}{S}}}] (s4) at (0,-8) {\rotatebox{-90}{\parbox[t][][r]{8mm}{\centering {\small 1-hot}}}};

    \node[neuronbox] (Hidd1) at (\layersep,-5) {\rotatebox{-90}{\parbox[t][][r]{8mm}{\centering {\small 32-d}}}};
    \node[neuronbox] (Hidd2) at (\layersep,-6) {\rotatebox{-90}{\parbox[t][][r]{8mm}{\centering {\small 32-d}}}};
    \node[neuronbox] (Hidd3) at (\layersep,-7) {\rotatebox{-90}{\parbox[t][][r]{8mm}{\centering {\small 32-d}}}};
    \node[neuronbox] (Hidd4) at (\layersep,-8) {\rotatebox{-90}{\parbox[t][][r]{8mm}{\centering {\small 32-d}}}};

    \draw[line width=5pt] (H1) -- (Hidd1);
    \draw[line width=5pt] (M2) -- (Hidd2);
    \draw[line width=5pt] (M3) -- (Hidd3);
    \draw[line width=5pt] (s4) -- (Hidd4); 

    \node[neuron,color=red] (F1-1) at (2*\layersep,-5.25) {};
    \node[neuron,color=red] (F1-2) at (2*\layersep,-6.25) {};
    \node[neuron,color=red] (F1-3) at (2*\layersep,-7.25) {};
    \node[neuron2] (F2-1) at (2*\layersep,-5.75) {};
    \node[neuron2] (F2-2) at (2*\layersep,-6.75) {};
    \node[neuron2] (F2-3) at (2*\layersep,-7.75) {};

    \draw[line width=1.5pt, color=red] (Hidd1) -- (F1-1);
    \draw[line width=1.5pt, color=red] (Hidd2) -- (F1-1);
    \draw[line width=1.5pt, color=red] (Hidd2) -- (F1-2);
    \draw[line width=1.5pt, color=red] (Hidd3) -- (F1-2);
    \draw[line width=1.5pt, color=red] (Hidd3) -- (F1-3);
    \draw[line width=1.5pt, color=red] (Hidd4) -- (F1-3);

    \draw[color=green,line width=1.5pt] (Hidd1) -- (F2-1);
    \draw[color=green,line width=1.5pt] (Hidd2) -- (F2-1);
    \draw[color=green,line width=1.5pt] (Hidd2) -- (F2-2);
    \draw[color=green,line width=1.5pt] (Hidd3) -- (F2-2);
    \draw[color=green,line width=1.5pt] (Hidd3) -- (F2-3);
    \draw[color=green,line width=1.5pt] (Hidd4) -- (F2-3);

    \node[neuron,color=red] (max-1) at (3*\layersep,-6.25) {};
    \node[neuron2] (max-2) at (3*\layersep,-6.75) {};

    \draw[->,line width=4pt,color=red] (F1-2) --++(0:1cm) node[above,midway]{};
    \draw[->,line width=4pt,color=green] (F2-2) --++(0:1cm) node[above,midway]{};
    
    \node[neuronbox] (Final) at (4*\layersep,-6.5){\rotatebox{-90}{\parbox[t][][r]{8mm}{\centering {\small 100-d}}}}; 

    \draw[line width=1.5pt] (max-1) -- (Final);
    \draw[line width=1.5pt] (max-2) -- (Final);
    




\end{tikzpicture}
  }%
  \caption{CNN. Each character is represented by a 1-hot vector, which is then mapped to a learned 32-d embedding vector. On these, $m$ ($m=2$ in the example) filters operate, which are combined to an $m$-dimensional vector via max-over-time-pooling. 
  The output layer, with $\tanh$ activation, is 100-d and is fully connected with the $m$-dim layer that feeds into it. We represent the left context, right context, and center via the same illustrated CNN, and then concatenate the 100-d representations to a 300-d representation of the input.
  }
  \label{fig:cnn}
\end{figure}
\subsection*{Deep Learning models}
Our first model is a {character-level convolutional neural network (\textbf{char-CNN})} illustrated in Figure~\ref{fig:cnn}.
This model (A) considers each of the three contexts (left, center, right) independently, representing them by a 100-dimensional vector as follows. Each character 
is represented by a 1-hot vector, which is then mapped to a 32-dimensional embedding 
(not pre-trained, and updated during learning).
Then $m$ filters, each of size $s$, are applied on the embedding layer.
Max-over-time pooling results in an $m$-dimensional layer which is fully connected with the 100-dimensional output layer, with $\tanh$ activation function. The 100-d representations of each context are then (B) concatenated, resulting in a 300-dimensional representation of the input. A final softmax layer predicts one of our three target classes. The hyper-parameters of this model---additional to $\ell,r,c$ mentioned above---are: number of filters $m$, filter size $s$, and a few others, such as the number of characters to consider in each context window
.

\begin{figure}[!b]
  \centering
  \resizebox{0.48\textwidth}{!}{%
    \newcommand{\stack}[5]{
  \foreach \i in {1,...,#1} {
    \draw[fill=white] #2 ++({-0.1*(#1)},{0.1*(#1)}) ++({0.1*\i},{-0.1*\i}) rectangle +#3;
  }
  \path #2 -- +#3 node[draw=none,rect,pos=0.5] (#5) {#4};
}

\begin{tikzpicture}[->,>=stealth',node distance=1cm,thick,
  every node/.style={circle,draw,minimum size=30,align=center,very thick},
  circ/.style={circle,minimum size=20,align=center,very thick,text width=0.6cm},
  rect/.style={rectangle,minimum size=10},
  inv/.style={draw=none}]

\node [inv,text width=0.3cm,minimum size=20] (x0) at (-1,0) {$...$};
\node [inv,text width=0.3cm,minimum size=20] (x0) at (5.2,0) {$...$};

\node [inv,text width=0.3cm,minimum size=20] (x1) at (0,0) {$x_{t-1}$};
\node [circ] (h11) at (0,2) {$h_{t-1}$};
\node [circ] (h21) at (0,4) {$h_{t-1}$};
\node [rect] (c1) at (0,6) {concat};
\node [inv,text width=1.5cm] (h10) at (-2.4,2) {forward};
\node [inv,text width=1.5cm] (h20) at (-2.4,4) {backward};

\draw  (x1) edge (h11);
\draw (h10) edge (h11);
\draw (x1) edge[bend angle=40, bend left] (h21);
\draw (h21) edge (h20);
\draw (h11) edge[bend angle=40, bend right] (c1);
\draw  (h21) edge (c1);

\node [inv,text width=0.3cm,minimum size=20] (x2) at (2,0) {$x_{t}$};
\node [circ] (h12) at (2,2) {$h_{t}$};
\node [circ] (h22) at (2,4) {$h_{t}$};
\node [rect] (c2) at (2,6) {concat};
\draw  (x2) edge (h12);
\draw (h11) edge (h12);
\draw (x2) edge[bend angle=40, bend left] (h22);
\draw (h22) edge (h21);
\draw (h12) edge[bend angle=40, bend right] (c2);
\draw  (h22) edge (c2);

\node [inv,text width=0.3cm,minimum size=20] (x3) at (4,0) {$x_{t+1}$};
\node [circ] (h13) at (4,2) {$h_{t+1}$};
\node [circ] (h23) at (4,4) {$h_{t+1}$};
\node [rect] (c3) at (4,6) {concat};
\node [rect,inv] (h14) at (6,2) {...};
\node [rect,inv] (h24) at (6,4) {...};
\draw  (x3) edge (h13);
\draw (h12) edge (h13);
\draw (x3) edge[bend angle=40, bend left] (h23);
\draw (h23) edge (h22);
\draw (h13) edge[bend angle=40, bend right] (c3);
\draw  (h23) edge (c3);
\draw (h13) edge (h14);
\draw (h24) edge (h23);

\node [rect] (softmax) at (2,8) {softmax};
\draw (c1) edge (softmax);
\draw (c2) edge (softmax);
\draw (c3) edge (softmax);

\node [rect] (at1) at (2,-4) {concat};
\node [rect] (at2) at (2,-5.4) {max-over-time pooling};
\draw (at1) edge (at2);

\stack{4}{(-1,-2.9)}{(6,1.2)}{convolutional layers\\filter widths = 2, 3, 5, 7}{filters};
\draw ([xshift=-45pt]filters.center) edge (at1);
\draw ([xshift=-15pt]filters.center) edge (at1);
\draw ([xshift=15pt]filters.center) edge (at1);
\draw ([xshift=45pt]filters.center) edge (at1);
\stack{4}{(-1,-2.9)}{(6,1.2)}{convolutional layers\\filter widths = 2, 3, 5, 7}{filters};

\node [minimum size=27,inv] (inv1) at (0,-1.9) {};
\draw (x1) edge (inv1);
\node [minimum size=27,inv] (inv2) at (2,-1.9) {};
\draw (x2) edge (inv2);
\node [minimum size=27,inv] (inv3) at (4,-1.9) {};
\draw (x3) edge (inv3);

\coordinate [rect,inv,minimum size=0,inner sep=0,outer sep=0,text width=0] (athelp) at (-2,3) {};
\draw [-] (at2) edge (-3.5,-5.4);
\draw [-] (-3.5,-5.4) edge (-3.5,3);
\draw [-] (-3.5,3) edge (-2,3);

\draw (athelp) edge[bend angle=10, bend left] (h11);
\draw (athelp) edge[bend angle=10, bend left] (h12);
\draw (athelp) edge[bend angle=10, bend left] (h13);
\draw (athelp) edge[bend angle=10, bend right] (h21);
\draw (athelp) edge[bend angle=10, bend right] (h22);
\draw (athelp) edge[bend angle=10, bend right] (h23);

\end{tikzpicture}
  }%
  \caption{Bi-LSTM with attention.  
Pre-trained word embeddings $x_{t}$ are fed to an ensemble of CNN layers with 4 different filter widths. For each timestep the outputs are concatenated and we employ max-over-time pooling.
The resulting attention vector is supplied to the nodes in the forward and backward LSTM layers. The output of both LSTM layers is concatenated to a 128-dim vector, which is fed to the final softmax layer.}
  \label{fig:bilstm}
\end{figure}
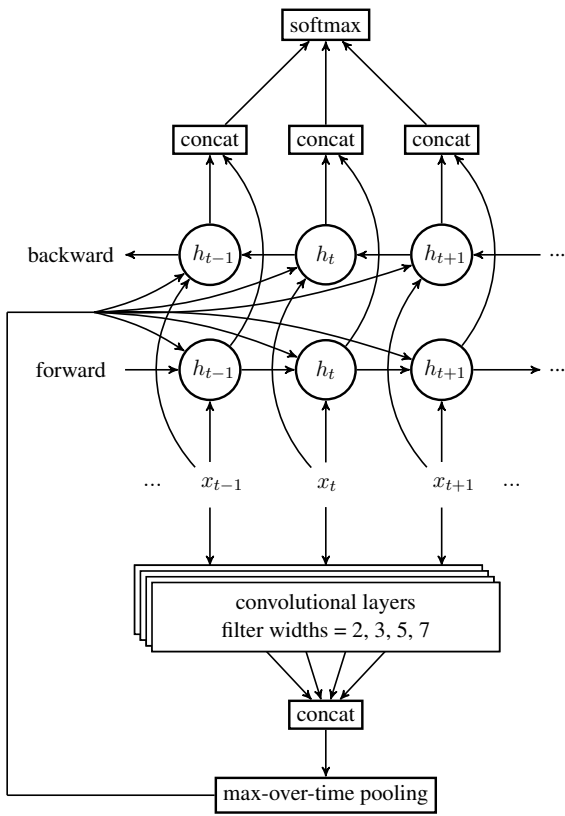

Our second model, which operates on the token-level, is a ``\textbf{stacked learner}''. We take five \emph{base classifiers} from scikit-learn (RandomForestClassifier with two different parameterizations; ExtraTreesClassifier with two different parameterizations; and XGBClassifier), and train them repeatedly on 90\% of the training data, extracting their predictions on the remaining 10\%. This process is iterated 10 times, in a cross-validation manner, so that we have a complete sample of predictions of the base classifiers on the training data. We then use a multi-layer perceptron (MLP) as a \emph{meta-classifier} that is trained to combine the predictions of the base classifiers into a final output prediction. The MLP is trained for 100 epochs and the model with best performance on a 10\% development set is chosen as the model to apply to unseen test data. 

Our third model (Figure~\ref{fig:bilstm}), also operating on the token level, is an attention based Bi-directional Long Short-Term Memory network (\textbf{AB-LSTM})\footnote{~Code was adapted from \url{https://github.com/codekansas/keras-language-modeling}}.
After loading pre-trained word embeddings, we apply 4 convolutional layers with filter sizes 2, 3, 5 and 7, followed by max-over-time-pooling. We concatenate the respective 
vectors to create an \emph{attention vector}.
The forward and backward LSTM layers (64-dimensional) are supplied with the pre-trained embeddings and 
the computed attention vector.
Their output is concatenated and, after applying dropout of 0.5, is used by the final softmax layer to predict 
the label probabilities.
\label{sec:approaches}
\section{Submitted Systems}

We set the $c$ hyper-parameter to $4$, and draw left and right context length hyper-parameters $\ell,r$ ($\ell=r$) from a discrete uniform distribution over the multi-set 
$\{1,2,2,3,3,3,4,4,4,4,5\}$.


Performance measure was micro-$F_{1}$ as computed by the task's evaluation script.\footnote{~We report results without the ``rel'' flag, i.e., corresponding to the column ``Overall'' in \citet{augenstein2017scienceie}, Table~4. Setting ``rel'' leads to consistently higher results. E.g., with this flag, we have 72\% micro-F$_1$ for our best ensemble (corresponding to column ``B'' in \citet{augenstein2017scienceie}, Table 4), rather than 69\% as reported in our Table~\ref{table:results}.} 
Table~\ref{table:results} shows average, maximum, and minimum performances of the systems we experimented with. We indicate the `incorrect' systems (those trained on only the dev set) with a star. 
We tested 56 different CNNs---hyper-parameters randomly drawn from specific ranges; e.g., we draw the number of filters $m$ from a normal distribution $\mathcal{N}(\mu=250,\sigma=50)$---
90 different stackers, and 20 different AB-LSTMs. Our three submitted systems were simple majority votes of (1) the 90 stackers, (2) the 90 stackers and 56 CNNs, (3) the 90 stackers, 56 CNNs and 20 AB-LSTMs. Overall, majority voting is considerably better than the mean performances of each system. 


\begin{table}[!htb]
\centering
\begin{tabular}{lrrr} \hline
\toprule
    & Mean & Max & Min \\ 
\midrule
CNN & 58.32$^*$/64.08 & 61$^*$/65 & 54$^*$/60 \\
Stacker & 61.57$^*$/67.11 &  64$^*$/68 & 59$^*$/65 \\
AB-LSTM & 59.12 & 64 & 56 \\
\midrule
Majority & 63$^*$/69 & 63$^*$/69 & 62$^*$/68 \\
\bottomrule
\end{tabular}
\caption{Micro-$F_{1}$ results in \% for our systems.}
\label{table:results}
\end{table}


For the stacker, the 
Komninos embeddings worked consistently best, with an average $F_{1}$-score of 63.83\%. Levy embeddings were second (62.50), followed by Glove embeddings of size 50 (61\%), size 300 (60.80) and size 100 (59.50).
We assume this is due to the Komninos embeddings being `richest' in nature, capturing both semantic and syntactic information. However, with more training data (corrected results), mean performances as a function of embedding type are closer: 67.77 (Komninos), 67.61 (Levy), 67.38 (Glove-300), 66.88 (Glove-50), 65.77 (Glove-100). 
  The AB-LSTM could not capitalize as much on the syntactic information, and performed best with the Glove embeddings, size 100 (60.35\%), and worst with the Levy embeddings (57.80).
  
The char-level CNN and the stacker performed individually considerably better than the AB-LSTM. However, including the AB-LSTM in the ensemble slightly increased the majority $F_1$-score on both the M and T class, as Table~\ref{table:maj} shows.
\begin{table}[!htb]
\centering
\begin{tabular}{lrrr} \hline
\toprule
  Ensemble  & M & P & T \\ 
\midrule
(1) {\small Stackers} & 76 & 71 & 46 \\
 (2) {\small Stackers$+$CNNs} & 76 &  72 & 46 \\
(3) {\small Stackers$+$CNNs$+$AB-LSTMs} & 77 & 72 & 47 \\
\bottomrule
\end{tabular}
\caption{$F_{1}$ results in \% across different classes.}
\label{table:maj}
\end{table}

\textbf{Error analysis}: Table~\ref{table:ab-lstm-confusion} details that {\textit{TASK} is often confused 
with \textit{PROCESS}, and---though less often---vice versa, leading to drastically lower $F_{1}$-score than for the other two classes. 
  This mismatch is 
  because 
  \textit{PROCESS} and \textit{TASK} can describe similar concepts, resulting in rather subtle differences. 
  E.g., looking at various `analysis' instances, we find that some are labeled as \textit{PROCESS} and others as \textit{TASK} in the gold data.  This holds even for a few seemingly very similar keyphrases (`XRD analysis', `FACS analysis'). The 
  ensemble has trouble 
  labeling this correctly, tagging 6 of 17 `analysis' instances wrongly. 
  Beyond further suspicious labelings in the data (e.g., `nuclear fissions reactors' as Task), other cases could have been resolved by knowledge of syntax (`anionic \emph{polymerization} of styrene' is a process, not a material) and/or POS tags, and by knowledge of common abbreviations such as `PSD'.}

We note that our submitted systems have the best $F_{1}$-score for the minority class \emph{TASK} (45\%$^*$/47\% vs.\ $\le$28\% for all other participants). Thus, our submission would have scored 1st using \emph{macro}-$F_{1}$ (60.66$^*$/65.33 vs.\ $\le$56.66), even in the erroneous setting of much less training data.

\begin{table}[!htb]
  \centering
  \begin{tabular}{llccc}
    \toprule
    \multicolumn{2}{c}{} & \multicolumn{3}{c}{Prediction}\\
    \multicolumn{2}{c}{} & Material & Process & Task\\\cmidrule{2-5}
    \multirow{3}{*}{\rotatebox{90}{Gold}} & Material & 710 & 194 & 0\\
    & Process & 218 & 708 & 28\\
    & Task & 22 & 105 & 67\\
    \bottomrule
  \end{tabular}
  \caption{Stackers$+$CNNs$+$AB-LSTMs confusion matrix.}
  \label{table:ab-lstm-confusion}
\end{table}  


\label{sec:submitted}
\section{Conclusion}
We present an ensemble-based keyphrase classification system 
which has 
achieved close-to-the-best results 
in the ScienceIE Subtask (B) while using only a fraction of the available training data. With the full training data, our approach ranks 1st. 
To avoid using expert features has been one of our priorities, but we believe that incorporating additional task-neutral information beyond words and word order would benefit the system performance.

We also experimented with document embeddings, created from additionally crawled ScienceDirect\footnote{~\url{https://dev.elsevier.com/api_docs.html}} articles. Even though the stacker described in \S\ref{sec:approaches} acting as a document classifier obtained a reasonably high accuracy of $\sim$87\%, its predictions had little effect on the overall results.

Manual examination of system errors shows that using part-of-speech tags, syntactic relations and simple named entity recognition would very likely boost the performance of our systems. 


\section*{Acknowledgments}
This work has been supported by the Volkswagen Foundation, FAZIT,  
DIPF,
KDSL,
and the EU's Horizon 2020 research and innovation programme (H2020-EINFRA-2014-2) under grant agreement \ftextnumero~654021. It reflects only the authors’ views and the EU is not liable for any use that may be made of the information contained therein.

\bibliography{semeval2017}
\bibliographystyle{acl_natbib}

\end{document}